\documentclass[runningheads]{llncs}
\usepackage{graphicx}
\usepackage{amsmath}
\usepackage{upgreek}
\usepackage{multirow}
\usepackage{multicol}
\usepackage{makecell}
\usepackage{booktabs}
\usepackage{subcaption}
\usepackage{xcolor}
\usepackage{wasysym}
\usepackage[T1]{fontenc}
\usepackage{float}
\begin{document}
\title{Detecting Outliers with Poisson Image Interpolation}
%
%
\author{
Jeremy Tan\inst{1} \and
Benjamin Hou\inst{1} \and
Thomas Day\inst{2} \and
John Simpson\inst{2} \and
Daniel Rueckert\inst{1} \and
Bernhard Kainz\inst{1,3}
}
%
\authorrunning{J. Tan et al.}
%
\institute{Imperial College London, SW7 2AZ, London, UK \and
King's College London, St Thomas' Hospital, SE1 7EH, London, UK \and Friedrich--Alexander University Erlangen--N\"urnberg, DE\\
\email{j.tan17@imperial.ac.uk}}
\maketitle              
\begin{abstract}
Supervised learning of every possible pathology is unrealistic for many primary care applications like health screening. Image anomaly detection methods that learn normal appearance from only healthy data have shown promising results recently. We propose an alternative to image reconstruction-based and image embedding-based methods and propose a new self-supervised method to tackle pathological anomaly detection. Our approach originates in the foreign patch interpolation (FPI) strategy that has shown superior performance on brain MRI and abdominal CT data. We propose to use a better patch interpolation strategy, Poisson image interpolation (PII), which makes our method suitable for applications in challenging data regimes. PII outperforms state-of-the-art methods by a good margin when tested on surrogate tasks like identifying common lung anomalies in chest X-rays or hypo-plastic left heart syndrome in prenatal, fetal cardiac ultrasound images. Code available at https://github.com/jemtan/PII.


\end{abstract}
\section{Introduction}
Doctors such as radiologists and cardiologists, along with allied imaging specialists such as sonographers shoulder the heavy responsibility of making complex diagnoses. Their decisions often determine patient treatment. Unfortunately, diagnostic errors lead to death or disability almost twice as often as any other medical error~\cite{tehrani201325}. In spite of this, the medical imaging workload has continued to increase over the last 15 years~\cite{bruls2020workload}. For instance, on-call radiology, which can involve high-stress and time-sensitive emergency scenarios, has seen a 4-fold increase in workload~\cite{bruls2020workload}. 

One of the major goals of anomaly detection in medical images is to find abnormalities that radiologists might miss due to excessive workload or inattention blindness~\cite{Drew2013Gorilla}. Most of the existing, automated methods are only suitable for detecting gross differences that are highly visible, even to observers without medical training. This undermines their usefulness in routine applications. Detecting anomalies at the level of medical experts typically requires supervised learning. This has been achieved for specific applications such as breast cancer~\cite{wu2019deep} or retinal disease~\cite{de2018clinically}. However, detecting arbitrary irregularities, without having any predefined target classes, remains an unsolved problem.   

Recently, self-supervised methods have proven effective for unsupervised learning~\cite{oord2018representation,henaff2019data}. Some of these methods use a self-supervised task that closely approximates the target task~\cite{chen2020simple,he2020momentum} (albeit without labels). There are also self-supervised methods that closely approximate the task of outlier detection. For example, foreign patch interpolation (FPI) trains a model to detect foreign patterns in an image~\cite{tan2020detecting}. The self-supervised task used for training takes a patch from one sample and inserts it into another sample by linearly interpolating pixel intensities. This creates training samples with irregularities that range from subtle to more pronounced. But for data with poor alignment and varying brightness, FPI's linear interpolation will lead to patches that are clearly incongruous with the rest of the image. This makes the self-supervised task too easy and reduces the usefulness of the learned features.

\noindent\textbf{Contribution:} We propose Poisson image interpolation (PII), a self-supervised method that trains a model to detect subtle irregularities introduced via Poisson image editing~\cite{perez2003poisson}. We demonstrate the usefulness of PII for anomaly detection in chest X-ray and fetal ultrasound data. Both of these are challenging datasets for conventional anomaly detection methods because the normal data has high variation and outliers are subtle in appearance.

\noindent\textbf{Related Work: }
Reconstruction-based outlier detection approaches can use autoencoders, variational autoencoders (VAEs)~\cite{zimmerer2019context}, adversarial autoencoders (AAEs)~\cite{chen2018unsupervised}, vector quantised variational autoencoders (VQ-VAE) \cite{razavi2019generating}, or generative adversarial networks (GANs)~\cite{schlegl2019f}. Some generative models are also used for pseudo-healthy image generation~\cite{xia2020pseudo}. Reconstruction can be performed at the image~\cite{baur2021autoencoders}, patch~\cite{wei2018anomaly}, or pixel~\cite{alaverdyan2020regularized} level. In each case, the goal is to replicate test samples as closely as possible using only features from the distribution of normal samples~\cite{baur2021autoencoders}. Abnormality is then measured as intensity differences between test samples and their reconstructions. Unfortunately, raw pixel differences lack specificity, making semantic distinctions more difficult.

Disease classifiers specialize in making fine semantic distinctions. They do this by learning very specific features and ignoring irrelevant variations~\cite{lecun2015deep}. 
To harness the qualities that make classifiers so successful, some methods compare samples as embeddings within a learned representation. For example, deep support vector data description (SVDD) learns to map normal samples to a compact hypersphere~\cite{ruff2018deep}. Abnormality is then measured as distance from the center of the hypersphere. Other methods, such as~\cite{marimont2021anomaly}, learn a latent representation using a VQ-VAE and exploit the autoregressive component to estimate the likelihood of a sample. Furthermore, components of the latent code with low likelihood can be replaced with samples from the learned prior. This helps to prevent the model from reconstructing anomalous features. A similar approach has also been proposed using transformers~\cite{pinaya2021unsupervised}. Overall, comparing samples in a learned representation space can allow for more semantic distinctions. But with only normal training examples, the learned representation may emphasize irrelevant features, \emph{i.e.}, those pertaining to variations \textit{within} the normal class. This often requires careful calibration for applications. 

Self-supervised methods aim to learn more relevant representations by training on proxy tasks. One of the most effective strategies is to train a classifier to recognize geometric transformations of normal samples~\cite{golan2018deep,tack2020csi}. By classifying transformations, the network learns prominent features that can act as reliable landmarks. Outliers that lack these key features will be harder to correctly classify when transformed. The anomaly score is thus inversely proportional to the classification accuracy. This works well for natural images, but in medical applications, disease appearance can be subtle. Many outliers still contain all of the major anatomical landmarks. 

To target more subtle abnormalities, some methods use a localized self-supervised task. For example, FPI synthesizes subtle defects within random patches in an image~\cite{tan2020detecting}. The corresponding pixel-level labels help the network to learn which regions are abnormal given the surrounding context. This approach showed good performance for spatially aligned brain MRI and abdominal CT data~\cite{mood_2020_3784230}. CutPaste~\cite{li2021cutpaste} also synthesizes defects by translating patches within an image. This is effective for detecting damage or manufacturing defects in natural images~\cite{li2021cutpaste}. However, unlike cracks or scratches seen in manufacturing, many medical anomalies do not have sharp discontinuities. Overfitting to obvious differences between the altered patch and its surroundings can limit generalization to more organic and subtle outliers. We propose to resolve this issue using Poisson image editing~\cite{perez2003poisson}. This helps to create more subtle defects (for training) which in turn improves generalization to real abnormalities.

\section{Method}
To begin, we provide a brief description of FPI. Consider two normal training samples, $x_i$ and $x_j$, of dimension $N\times N$, as well as a random patch $h$, and a random interpolation factor $\alpha \in [0,1]$. FPI replaces the pixels in patch $h$ with a convex combination of $x_i$ and $x_j$, to produce a training image $\widetilde{x}_i$ (Eqn.~\ref{equation:patch_interp}). Note that $\widetilde{x}_i=x_i$ outside of $h$. For a given training image $\widetilde{x}_i$, the corresponding label is $\widetilde{y}_i$, as specified by Eqn.~\ref{equation:label}. 

\begin{equation}
\widetilde{x}_{i_p}= (1-\alpha) x_{i_p} + \alpha x_{j_p} \;, \; \forall\; p \in h
\label{equation:patch_interp}
\end{equation}

\begin{equation}
\widetilde{y}_{i_p}= 
\begin{cases}
    \alpha  & \textrm{if } p \in h \\    
    0       & \textrm{otherwise}    
\end{cases}
\label{equation:label}
\end{equation}

This approach has similarities to mixup~\cite{zhang2017mixup}, a data augmentation method that generates convex combinations of images and their respective labels. In the case of FPI, the training data only contains normal samples. Without having any class labels, FPI calculates its own labels as convex combinations of self ($0 \textrm{ for } y_i$) and non-self ($1 \textrm{ for } y_j$) as shown in  Eqn.~\ref{equation:label}. 

\begin{figure*}[h]
	\centering
        \includegraphics[width=\linewidth]{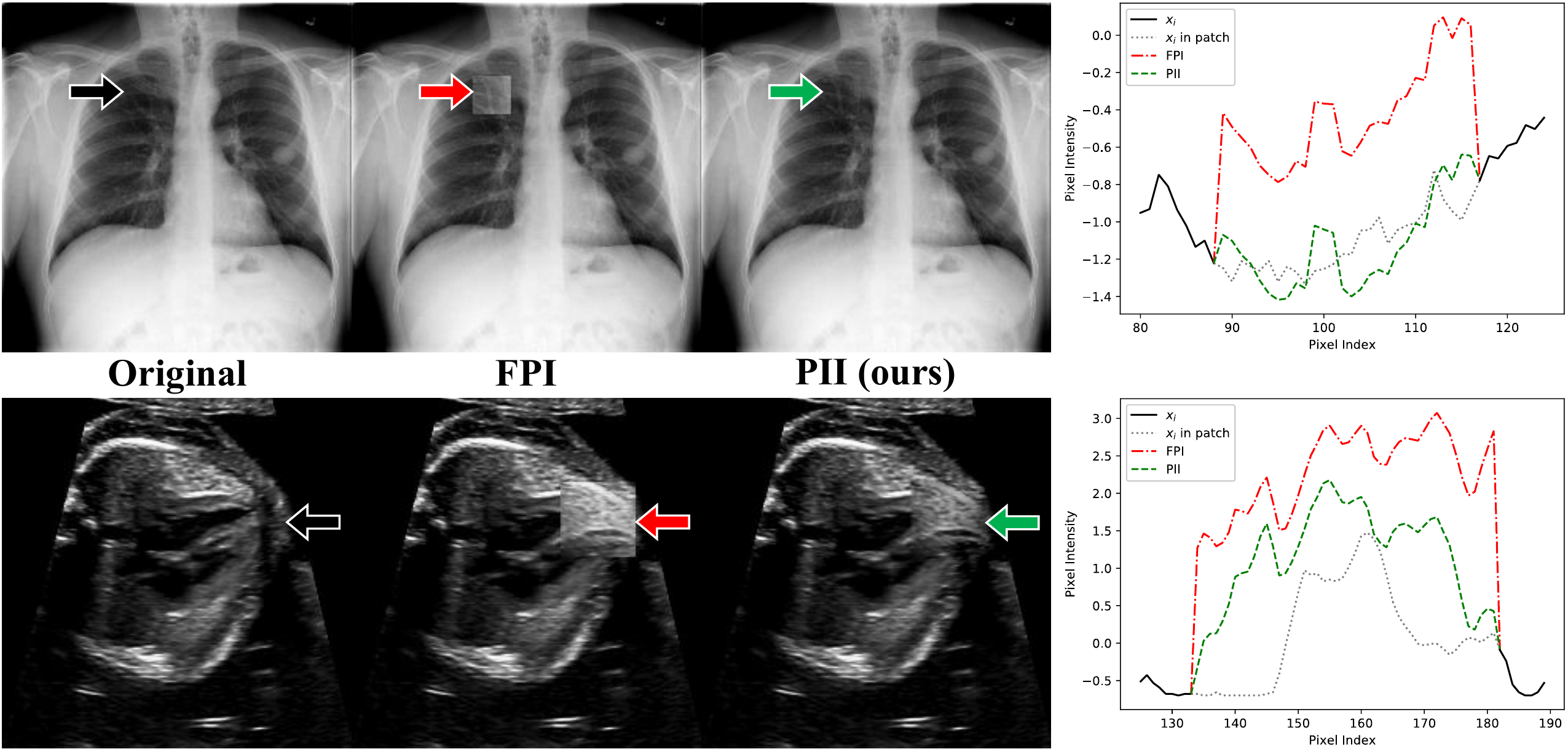}
		\caption{\textit{Examples of patches altered by FPI (convex combination) and PII (Poisson blending). Arrows in the images indicate the location of the line plotted on the right. Altering patches can simulate subtle (top) or dramatic (bottom) changes to anatomical structures. In both cases, PII blends the changes into the image more naturally. 
		}}
	\label{figures:Blending}
\end{figure*}

If $x_i$ and $x_j$ have vastly different intensity levels or structures, the interpolated patch will be inconsistent with the rest of the image. These differences are easy to spot and provide no incentive for the model to learn features that constitute ``normal'' (a much harder task). To create more challenging cases, we use a technique for seamless image blending. Poisson image editing~\cite{perez2003poisson} blends the content of a source image ($x_j$) into the context of a destination image ($x_i$). Rather than taking the raw intensity values from the source, we extract the relative intensity differences across the image, \emph{i.e.} the image gradient. Combining the gradient with Dirichlet boundary conditions (at the edge of the patch) makes it possible to calculate the absolute intensities within the patch. This is illustrated in Figure~\ref{figures:Blending}. 

More formally, let $f_{in}$ be a scalar function representing the intensity values within the patch $h$. 
The goal is to find intensity values of $f_{in}$ that will:
\begin{enumerate}
    \item match the surrounding values, $f_{out}$, of the destination image, along the border of the patch ($\partial h$), and 
    \item follow the relative changes (image gradient), $\textbf{v}$, of the source image.
\end{enumerate} 

\begin{equation}
\min_{f_{in}} \iint \limits_{h} \left| \nabla f_{in} - \textbf{v} \right| ^2 
\textrm{ with } \; \; f_{in} \Big|_{\partial h} =  f_{out} \Big|_{\partial h}
\label{equation:minf}
\end{equation}

\begin{equation}
\Updelta f_{in} =  \textrm{div}\textbf{v} \textrm{ over } h,
\textrm{ with } \; \; f_{in} \Big|_{\partial h} =  f_{out} \Big|_{\partial h}
\label{equation:poissonEqn}
\end{equation}

These conditions are specified in Eqn.~\ref{equation:minf}~\cite{perez2003poisson} and its solution is the Poisson equation (Eqn.~\ref{equation:poissonEqn}). 
Intuitively, the Laplacian, ($\Updelta\cdot = \frac{\partial^2 \cdot}{\partial x^2} + \frac{\partial^2 \cdot}{\partial y^2}$), should be close to zero in regions that vary smoothly and have a larger magnitude in areas where the gradient ($\textbf{v}$) changes quickly.  

To find $f_{in}$ for discrete pixels, a finite difference discretization can be used. Let $p$ represent a pixel in $h$ and let $q \in N_p$ represent the four directly adjacent neighbours of $p$. The solution should satisfy Eqn.~\ref{equation:discreteVersion} (or Eqn.~\ref{equation:discreteVersionBoundary} if any neighbouring pixels $q$ overlap with the patch boundary $\partial h$)~\cite{perez2003poisson}.

\begin{equation}
\left| N_p \right| f_{in_p} - \sum_{q \in N_p} f_{in_q}=  \sum_{q \in N_p} \textrm{v}_{pq}
\label{equation:discreteVersion}
\end{equation}

\begin{equation}
\sum_{q \in N_p \cap h} \left( f_{in_p} - f_{in_q} \right) = \sum_{q \in N_p \cap \partial h} f_{out_q} + \sum_{q \in N_p} \textrm{v}_{pq}
\label{equation:discreteVersionBoundary}
\end{equation}

In our case the image gradient comes from finite differences in the source image $x_j$, i.e. $\textrm{v}_{pq}=x_{j_p}-x_{j_q}$. Meanwhile, the boundary values, $f_{out_q}$, come directly from the destination image, $x_{i_q}$. This system can be solved for all $p \in h$ using an iterative solver. In some cases, this interpolation can cause a smearing effect. For example, when there is a large difference between the boundary values at opposite ends of the patch, but the gradient within the patch (from $x_j$) is very low. To compensate for this, Perez et al. suggest using the original gradient (from $x_i$) if it is larger than the gradient from $x_j$ (Eqn.~\ref{equation:gradientLarger})~\cite{perez2003poisson}. We modify this to introduce the interpolation factor, $\alpha$, that FPI uses to control the contribution of $x_j$ to the convex combination. In this case, $\alpha$ controls which image gradients take precedence (Eqn.~\ref{equation:gradientLargerAlpha}). This creates more variety in training samples, i.e. more ways in which two patches can be combined. It also helps create a self-supervised task with varying degrees of difficulty, ranging from very subtle to more prominent structural differences. Figure~\ref{figures:Blending} demonstrates that this formulation can blend patches seamlessly. 

\begin{equation}
\textrm{v}_{pq} = 
\begin{cases}
    x_{i_p}-x_{i_q} \textrm{ if } \left| x_{i_p}-x_{i_q} \right| > \left| x_{j_p}-x_{j_q} \right| \\
    x_{j_p}-x_{j_q} \textrm{ otherwise}
\end{cases}
\label{equation:gradientLarger}
\end{equation}

\begin{equation}
\textrm{v}_{pq} = 
\begin{cases}
    \left(1-\alpha\right) \left(x_{i_p}-x_{i_q}\right) 
    & \textrm{ if } \left|\left(1-\alpha\right) \left(x_{i_p}-x_{i_q}\right) \right| > 
    \left| \alpha \left(x_{j_p}-x_{j_q}\right) \right| \\
    \alpha \left(x_{j_p}-x_{j_q}\right) 
    & \textrm{ otherwise}
\end{cases}
\label{equation:gradientLargerAlpha}
\end{equation}

\begin{figure*}[h]
	\centering
        \includegraphics[width=\linewidth]{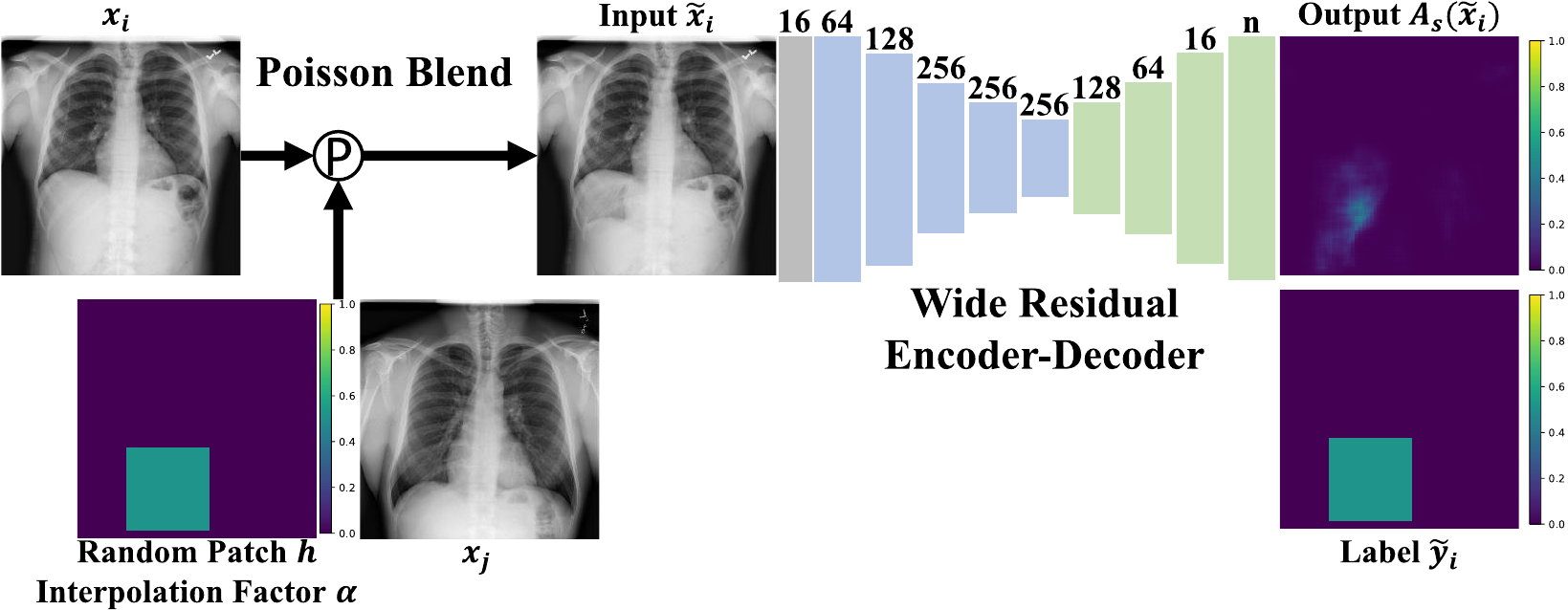}
		\caption{\textit{Illustration of PII self-supervised training. The network architecture starts with a single convolutional layer (gray), followed by residual blocks (blue/green). Values above each block indicate the number of feature channels in the convolutional layers. In all experiments we use a single output channel, i.e. $n=1$.}}
	\label{figures:architecture}
\end{figure*}

PII uses the same loss as FPI, which is essentially a pixel-wise regression of the interpolation factor $\alpha$~\cite{tan2020detecting}. The loss is given in Eqn.~\ref{equation:loss_regression}:
\begin{equation}
\mathcal{L}_{\textrm{bce}} = -\widetilde{y}_{i_p} logA_s(\widetilde{x}_{i_p}) -(1-\widetilde{y}_{i_p})\textrm{log}(1-A_s(\widetilde{x}_{i_p}))
\label{equation:loss_regression}
\end{equation}
The inputs, $\widetilde{x}_i$, are training samples that contain a random patch with values $f_in$, computed via Poisson blending as described above. The output of the model is used directly as an anomaly score, $A_s$. A diagram of the setup is given in Figure~\ref{figures:architecture}.

\noindent\textbf{Architecture and Specifications:}

We follow the same network architecture as FPI, a wide residual encoder-decoder~\cite{tan2020detecting}. The encoder is a standard wide residual network~\cite{Zagoruyko2016Wide} with a width of 4 and depth of 14. The decoder has the same structure but in reverse. The output has the same shape as the input and uses a sigmoid activation. Training is done using Adam~(\cite{kingma2014adam}) with a learning rate of $10^{-3}$ for 50 epochs. The self-supervised task uses patches, $h$, that are randomly generated with size $h_{s} \sim U(0.1N,0.4N)$, and center coordinates $h_{c} \sim U_{2}(0.1N,0.9N)$. Each patch is also given a random interpolation factor $\alpha \sim U(0.05,0.95)$.

\noindent\textbf{Implementation}: We use TensorFlowV1.15 and train on a Nvidia TITAN Xp GPU. Training on our largest dataset for 50 epochs takes about 11 hours. PII solves partial differential equations on the fly to generate training samples dynamically. To achieve this we use multiprocessing~\cite{mckerns2012building} to generate samples in parallel. The code is available at https://github.com/jemtan/PII. 

\section{Evaluation and Results}
To evaluate the performance of PII, we compare with an embedding-based method, a reconstruction-based method, and a self-supervised method. For the embedding-based method, we use Deep SVDD~\cite{ruff2018deep} with a 6 layer convolutional neural network. Meanwhile, a vector-quantized variational autoencoder  (VQ-VAE2)~\cite{razavi2019generating} is used as a reconstruction-based method. The VQ-VAE2 is trained using the same wide residual encoder-decoder architecture as PII, except the decoder is given the same capacity as the encoder to help produce better reconstructions. FPI~\cite{tan2020detecting}, which our method builds upon, is used as a self-supervised benchmark method. To compare each method, we calculate the average precision (AP) for each of the datasets described below. Average precision is a scalar metric for the area under the precision-recall curve.

\noindent\textbf{Data:}
Our first dataset is ChestX-ray14~\cite{wang2017chestx}, a public chest X-ray dataset with  108,948 images from 32,717 patients  showing 14 pathological classes as well as a normal class. From this large dataset, we extract 43,322 posteroanterior (PA) views of adult patients (over 18) and split them into male (\mars) and female (\female) partitions. All X-ray images are resized to 256x256 (down from 1024x1024) and normalized to have zero mean and unit standard deviation. The training/test split is summarized in Table~\ref{table:results}. 

The second dataset consists of a total of 13380 frames from 108 patients acquired during routine fetal ultrasound screening. This is an application where automated anomaly detection in screening services would be of most use. We use cardiac standard view planes~\cite{Fasp2018}, specifically 4-chamber heart (4CH) and 3-vessel and trachea (3VT) views, from a private and de-identified dataset of ultrasound videos. For each selected standard plane, 20 consecutive frames (10 before and 9 after) are extracted from the ultrasound videos. Images are 224x288 and are normalized to zero mean, unit standard deviation. Normal samples consist of healthy images from a single view (4CH/3VT) and anomalous images are composed of alternate views (3VT/4CH) as well as pathological hearts of the same view (4CH/3VT). For pathology we use cases of hypoplastic left heart syndrome (HLHS), a condition that affects the development of the left side of the heart~\cite{simpson2000hypoplastic}. The training/test split is outlined in Table~\ref{table:results}. The scans are of volunteers at 18-24 weeks gestation (Ethics: \emph{\emph{anonymous during review}}), in a fetal cardiology clinic, where patients are referred to from primary screening and secondary care sites. Video clips have been acquired on Toshiba Aplio  i700, i800 and Philips EPIQ V7 G devices.

\noindent\textbf{Results:}
We compare our method with recent state-of-the-art anomaly detection methods in Table~\ref{table:results}. 



\begin{table}[ht]
\centering 
\begin{tabular}{l c c c c c}
\toprule
\multirow{2}{*}{\textbf{Dataset}}
& \multicolumn{3}{c}{{Chest X-ray}} &  \multicolumn{2}{c}{{Fetal US}} \\
 \cmidrule(lr){2-4}
 \cmidrule(lr){5-6}
& \textbf{\mars PA} & \textbf{\female PA} & & \textbf{4CH} & \textbf{3VT} \\
\midrule
 & \multicolumn{5}{c}{Number of Images} \\
\cmidrule(lr){2-4} 
\cmidrule(lr){5-6}
Normal Train & 17852 & 14720 & & 283$\times$20 & 225$\times$20 \\
Normal Test & 2634 & 2002 & & 34$\times$20 & 35$\times$20 \\
Anomalous Test & 3366 & 2748 & & 54$\times$20 & 38$\times$20 \\
\midrule
 & \multicolumn{5}{c}{Average Precision} \\
 \cmidrule(lr){2-4} 
 \cmidrule(lr){5-6}
Deep SVDD & 0.565 & 0.556 &  & 0.685 & 0.893 \\
VQ-VAE2 & 0.503 & 0.516 &  & 0.617 & 0.578 \\
FPI & 0.533 & 0.586 &  & 0.658 & 0.710 \\
PII & \textbf{0.690} & \textbf{0.703} &  & \textbf{0.723} & \textbf{0.929} \\
\bottomrule
\end{tabular}
\caption{Each dataset is presented in one column. The train-test split is shown for each partition (top). Note that ultrasound images are extracted from videos as 20 frame clips. Average precision is also listed for each method (bottom).}
\label{table:results}
\end{table}

Example test images are shown in Figure~\ref{figures:cxrOutput}. The VQ-VAE2 reconstruction error indicates that sharp edges are difficult to reproduce accurately. 
Meanwhile FPI is sensitive to sharp edges because of the patch artifacts produced during training. In contrast, PII is sensitive to specific areas that appear unusual.  

\begin{figure*}[ht]
        \includegraphics[width=0.5\linewidth]{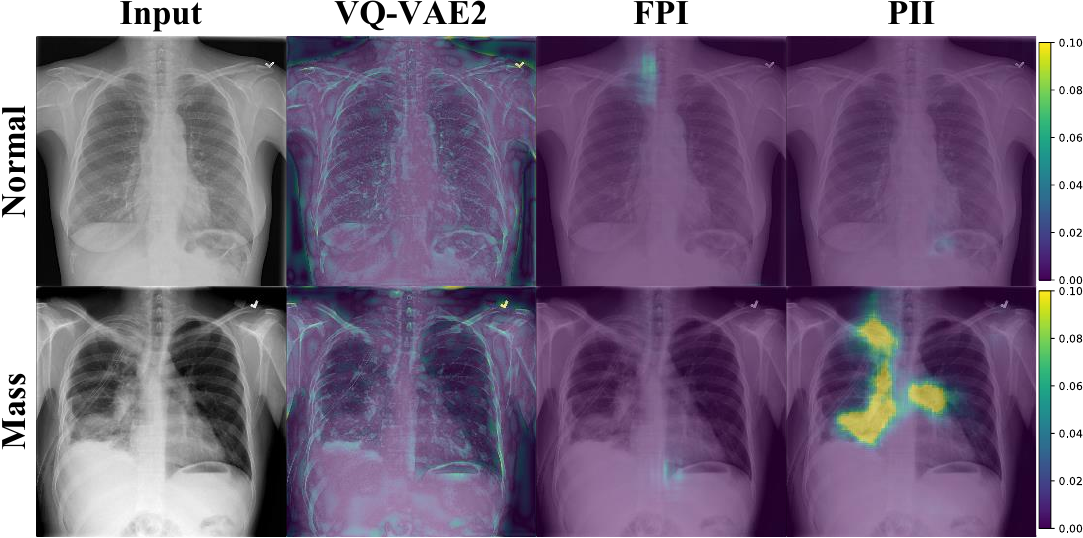}
        \includegraphics[width=0.5\linewidth]{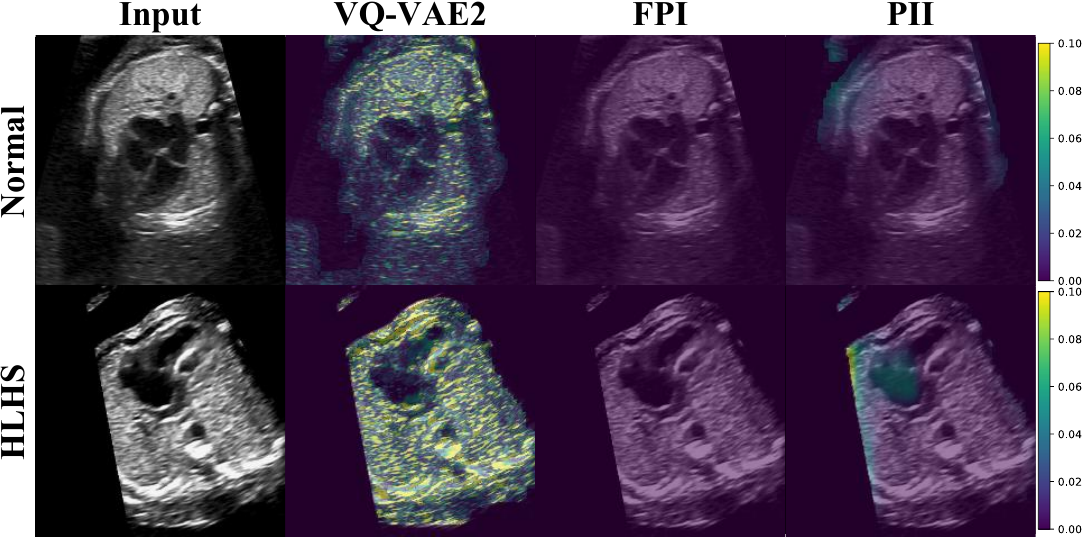}
\caption{\textit{Examples of test X-ray (left) and ultrasound (right) images with pixel-wise anomaly scores from each method. Note that the VQ-VAE2 reconstruction error is scaled down by a factor of 10.}}
\label{figures:cxrOutput}
\end{figure*}

\begin{figure}[H]
	\centering
	\begin{subfigure}[t]{0.32\linewidth}
        \includegraphics[width=\linewidth,keepaspectratio]{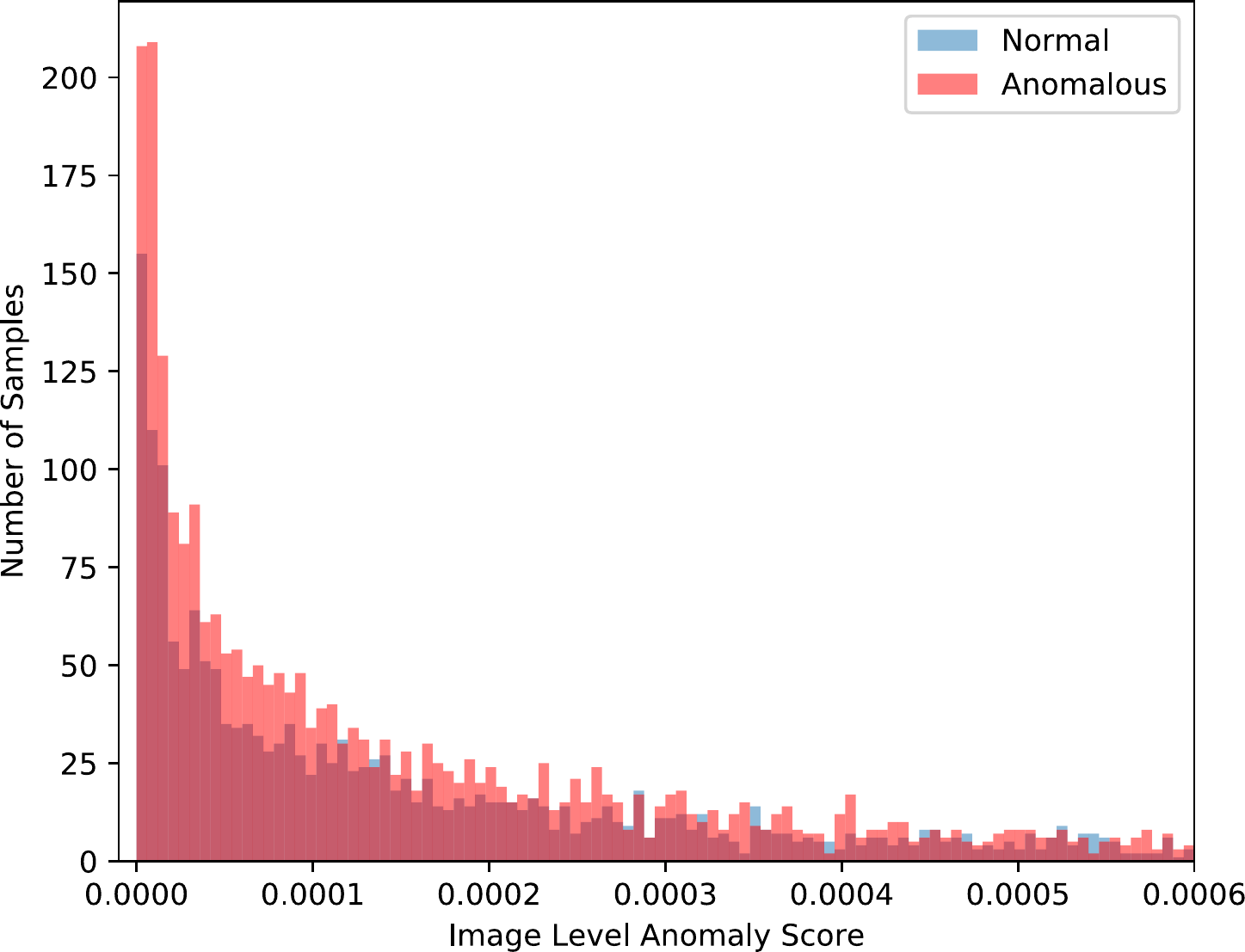}
	\end{subfigure}
	\hspace*{\fill}
	\begin{subfigure}[t]{0.32\linewidth}
		\includegraphics[width=\linewidth,keepaspectratio]{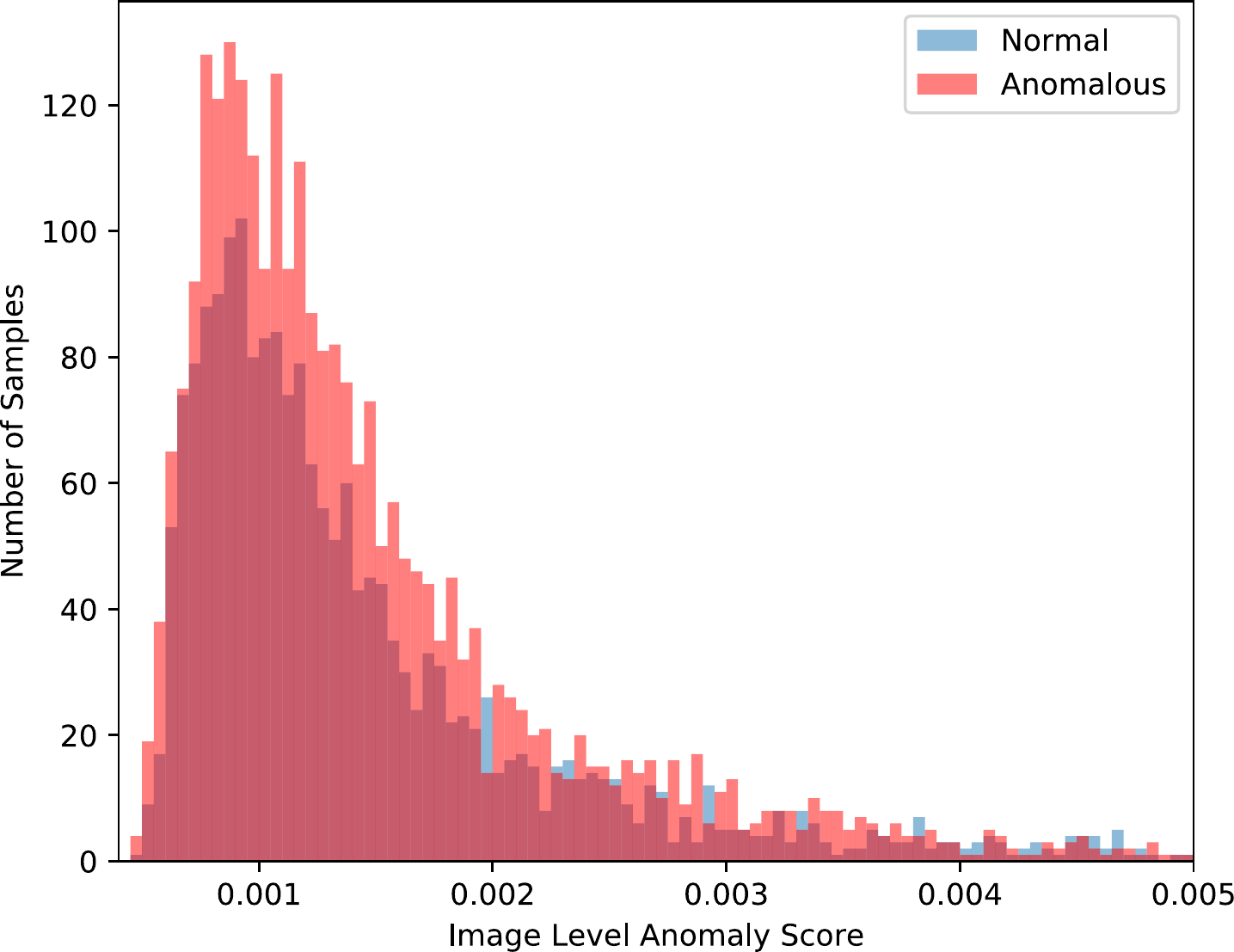}
	\end{subfigure}
	\hspace*{\fill}
	\begin{subfigure}[t]{0.32\linewidth}
		\includegraphics[width=\linewidth,keepaspectratio]{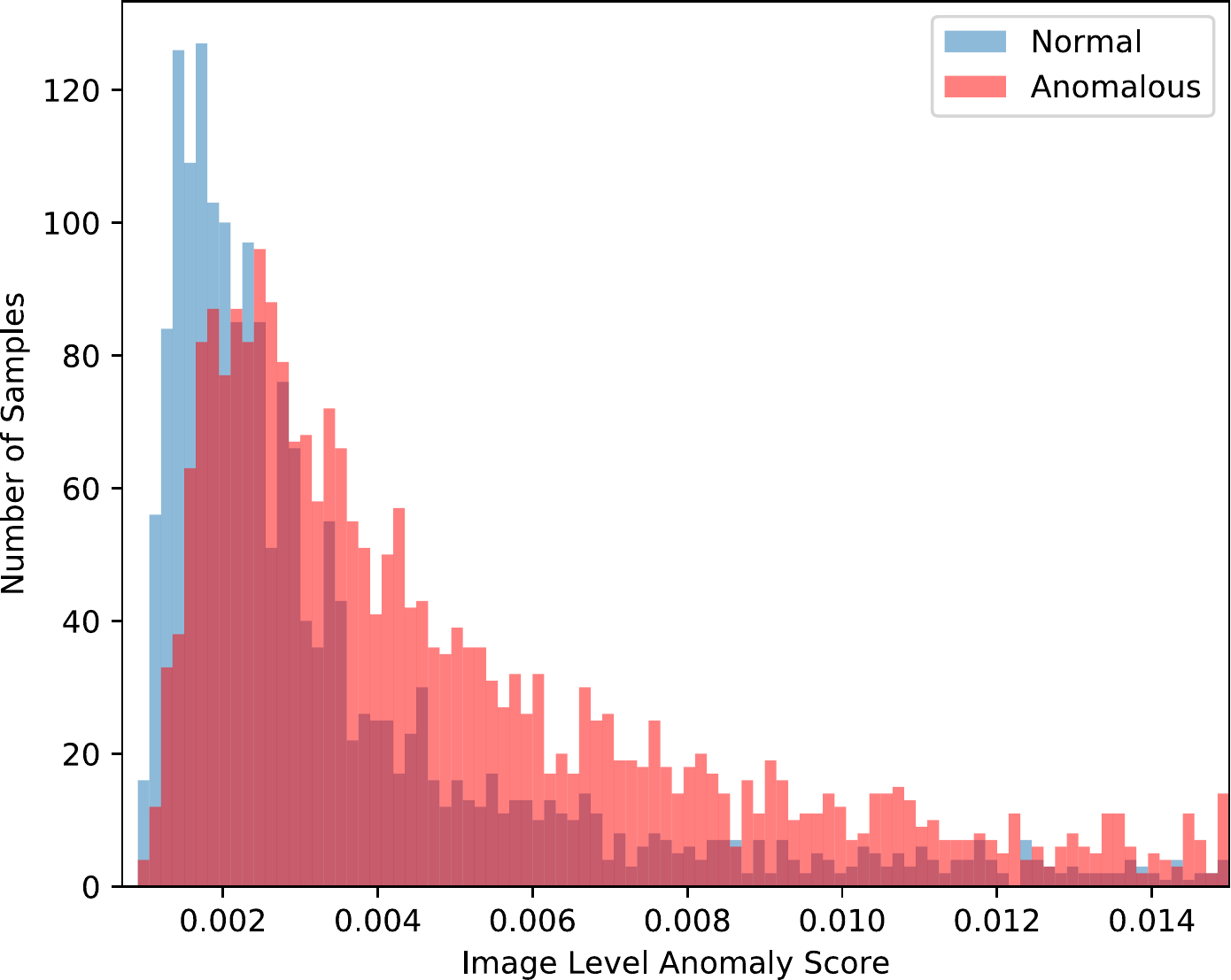}
	\end{subfigure}
	\hspace*{\fill}

	\begin{subfigure}[t]{0.32\linewidth}
        \includegraphics[width=\linewidth,keepaspectratio]{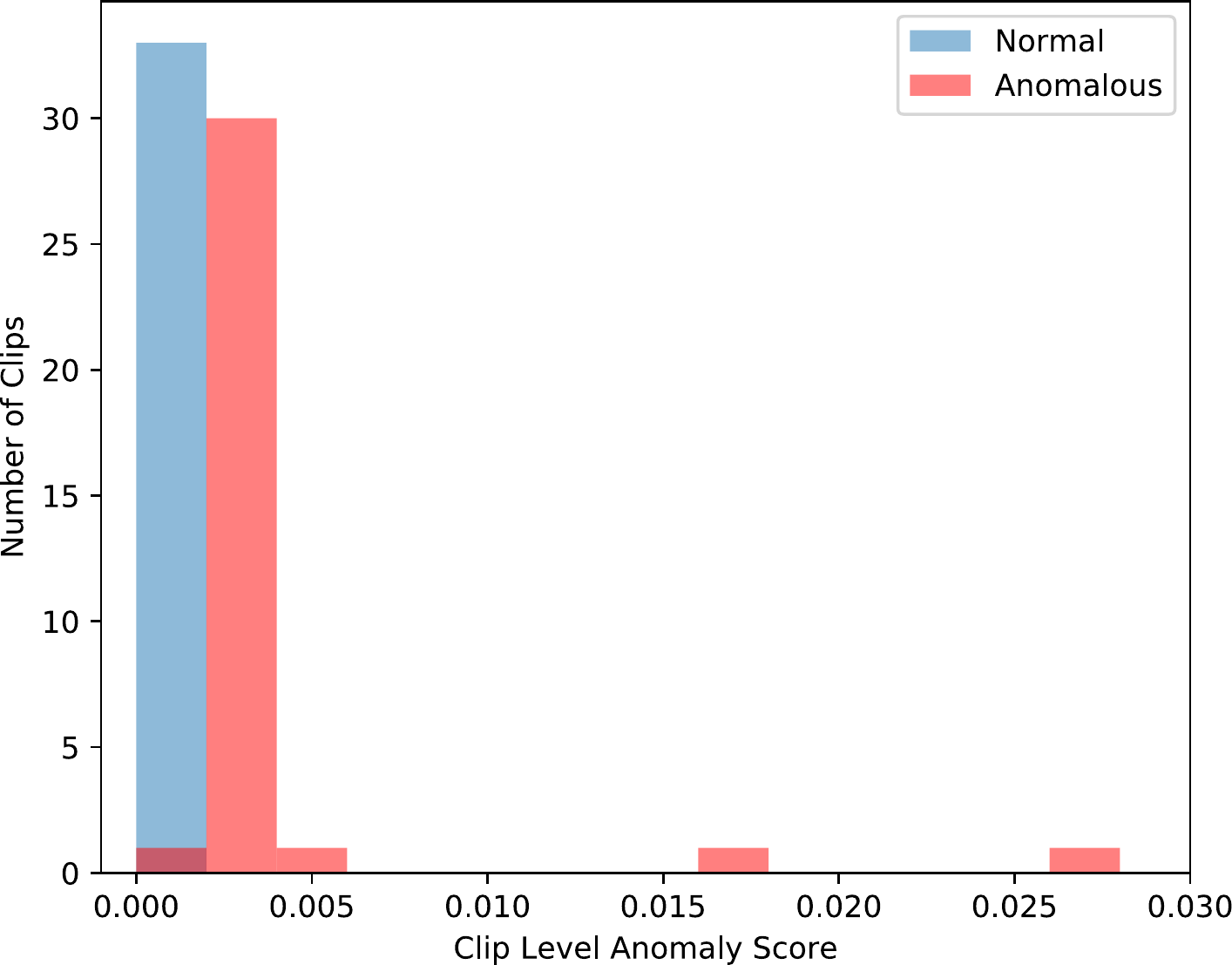}
		\caption{\textit{Deep SVDD}}
	\end{subfigure}
	\hspace*{\fill}
	\begin{subfigure}[t]{0.32\linewidth}
		\includegraphics[width=\linewidth,keepaspectratio]{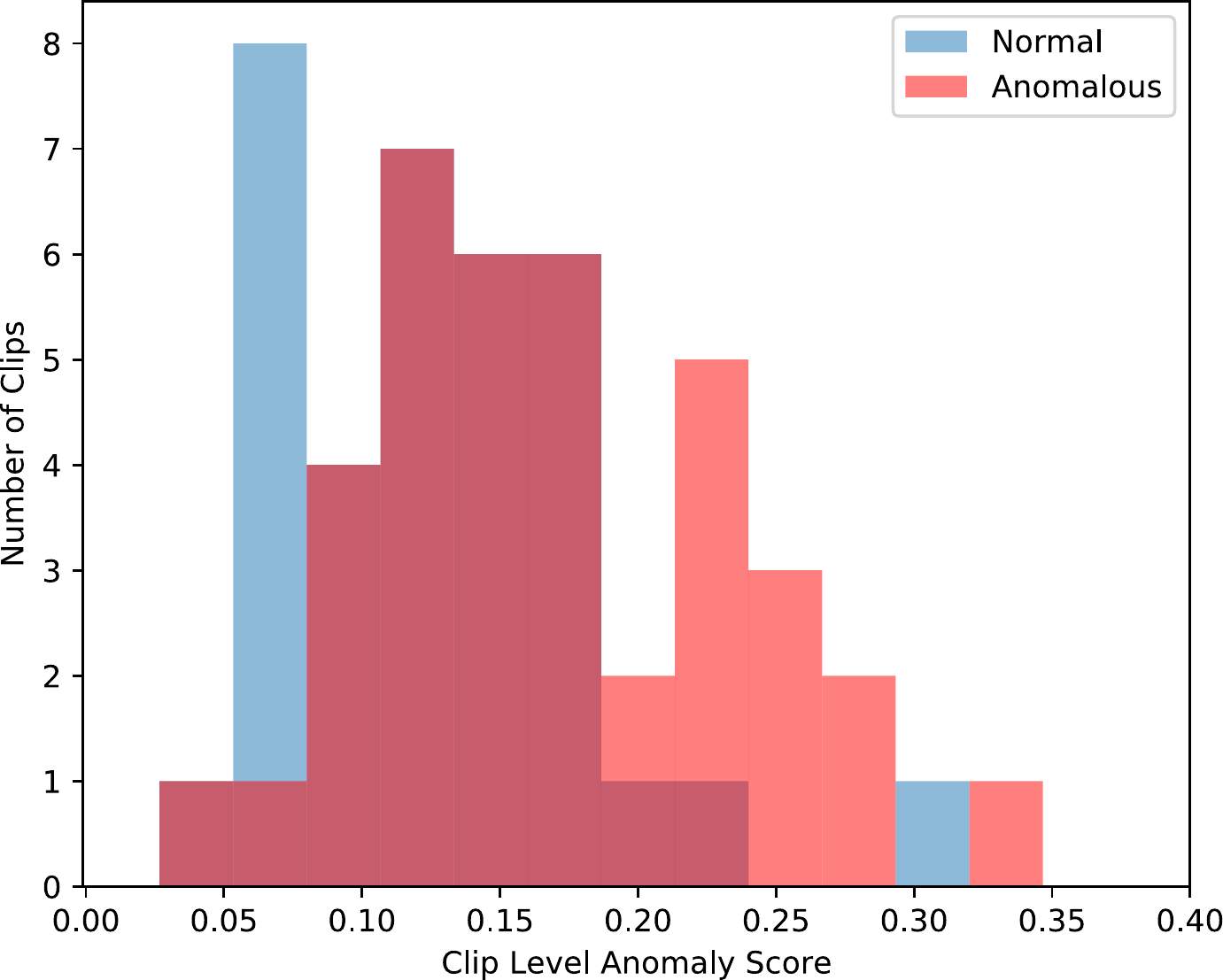}
		\caption{\textit{FPI}}
	\end{subfigure}
	\hspace*{\fill}
	\begin{subfigure}[t]{0.32\linewidth}
		\includegraphics[width=\linewidth,keepaspectratio]{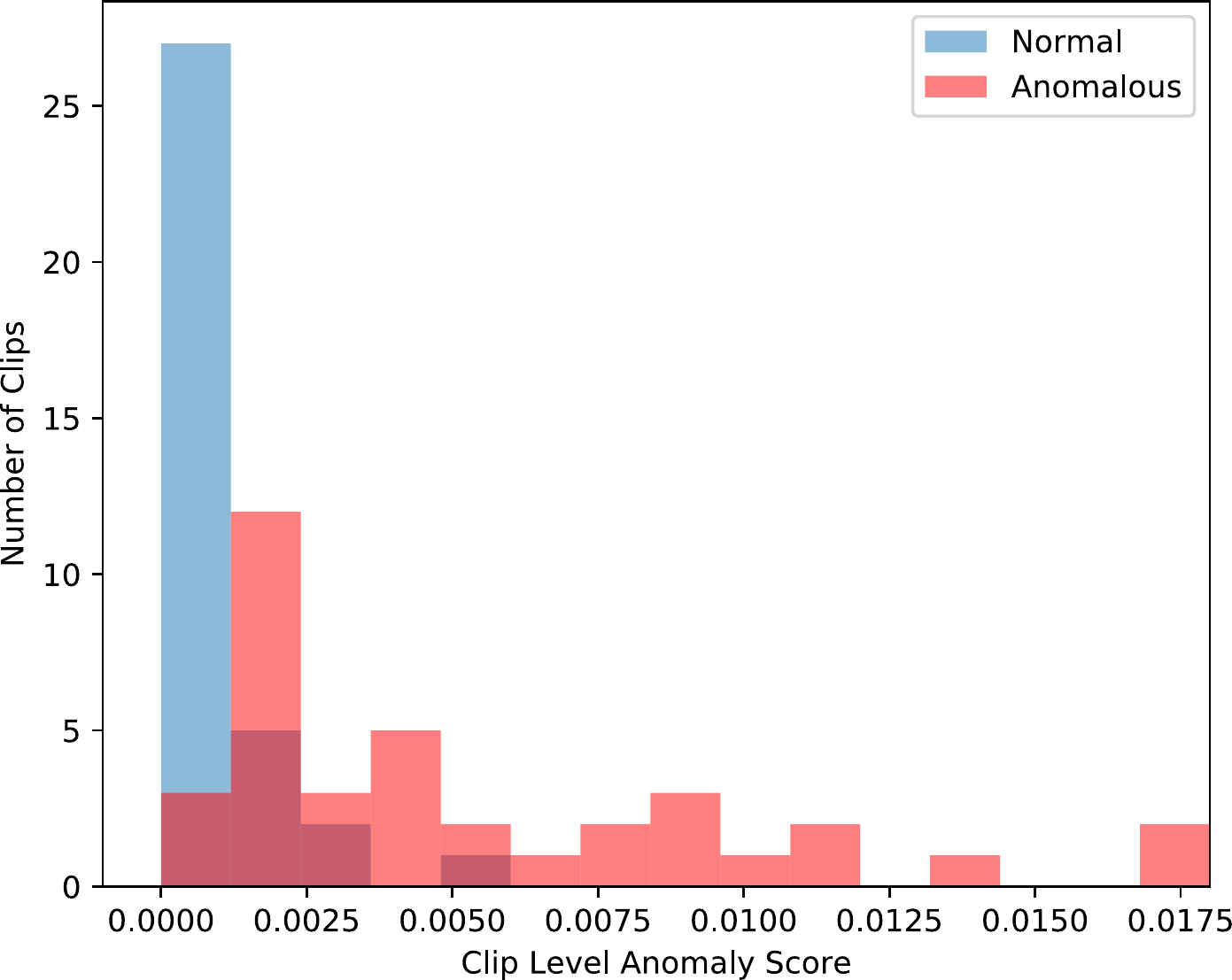}
		\caption{\textit{PII}}
	\end{subfigure}
	\caption{\textit{Histograms of image level anomaly scores for Chest X-ray Female PA data (top) and clip level anomaly scores for Fetal US 3VT data (bottom).}}
	\label{figures:histogram}		
\end{figure}

To see each method's ability to separate normal from anomalous, we plot histograms of anomaly scores for each method in Figure~\ref{figures:histogram}. The difficulty of these datasets is reflected in the fact that existing methods have almost no discriminative ability. On average, PII gives anomalous samples slightly higher scores than normal samples. 
The unusually high performance in the 3VT dataset is partly due to the small size of the dataset as seen in Figure~\ref{figures:histogram}. 

\noindent\textbf{Discussion:}
We have shown that our method is suitable to detect pathologies when they are considered anomalies compared to a training set that contains only healthy subjects. Training from only normal data is an important aspect in our field since a) data from healthy volunteers is usually available more easily, b) prevalence for certain conditions is low, thus collecting a well balanced training set is challenging and c) supervised methods would require in the ideal case equally many samples from every possible disease they are meant to detect. The latter is particularly a problem in rare diseases where the number of patients are very low in the global population. 


\section{Conclusion}
In this work we have discussed an alternative to reconstruction-based anomaly detection methods. We base our method on the recently introduced FPI method, which formulates a self-supervised task through patch-interpolation based on normal data only. We advance this idea by introducing Poisson Image Interpolation, which mitigates interpolation issues for challenging data like chest X-Rays and fetal ultrasound examinations of the cardio-vascular system. In future work we will explore spatio-temporal support for PII, which is in particular relevant for ultrasound imaging.  

\noindent\textbf{Acknowledgements:} Support from Wellcome Trust IEH Award iFind project [102431] and UK Research and Innovation London Medical Imaging and Artificial Intelligence Centre for Value Based Healthcare. JT was supported by the ICL President's Scholarship. 

%
%
%
%
\bibliographystyle{splncs04}
\bibliography{ref}

\clearpage

\appendix
\section{Additional Chest X-ray Samples}
\label{appendix:cxr}
\begin{figure*}[ht]
        \includegraphics[width=0.8\linewidth]{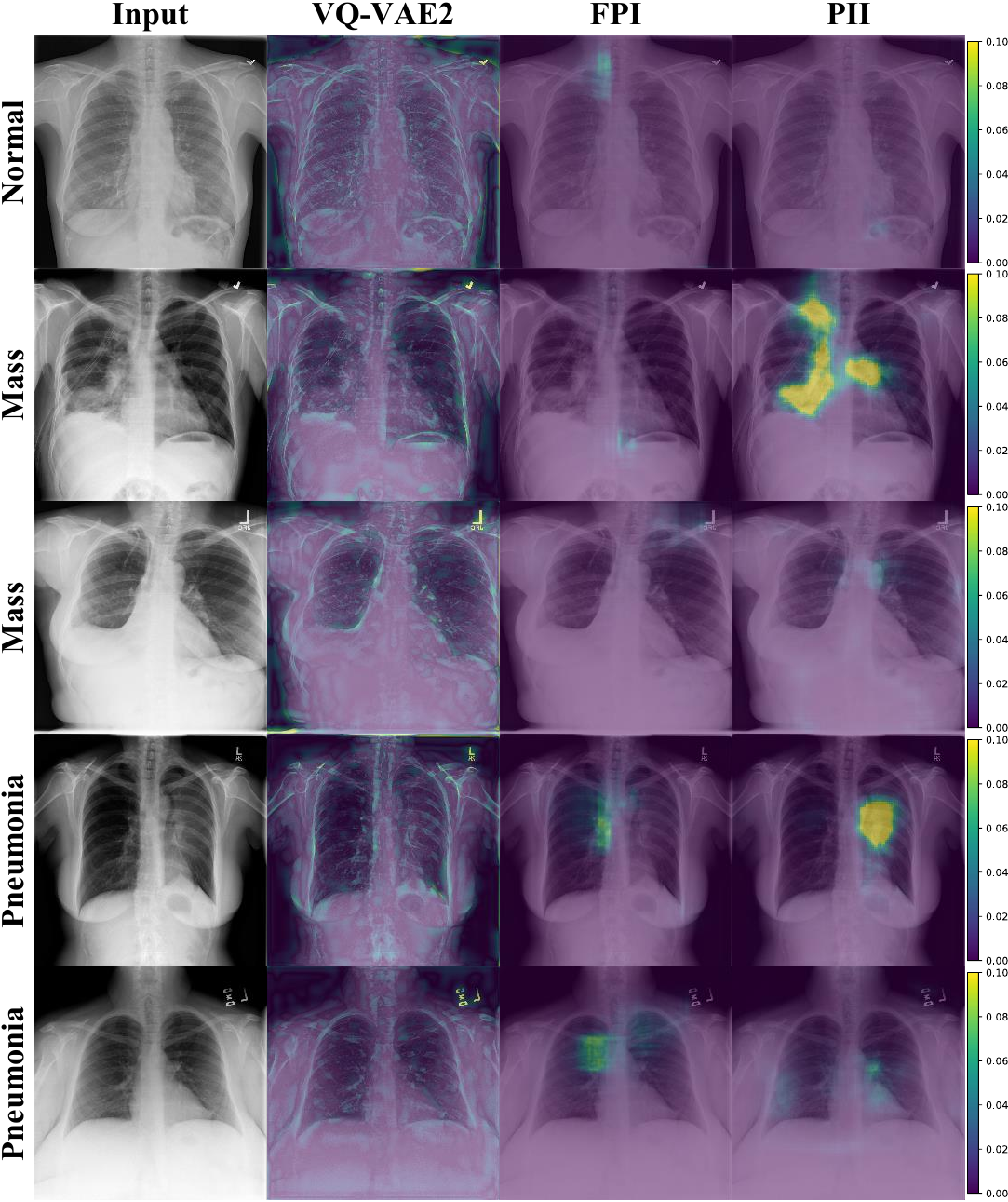}
\caption{\textit{Examples of test X-ray images with pixel-wise anomaly scores from each method. Note that the VQ-VAE2 reconstruction error is scaled down by a factor of 10.}}
\label{figures:cxrOutput}
\end{figure*}

\end{document}